# DebateSum:
# A large-scale argument mining and summarization dataset

**Allen Roush**
University of Oregon
gedboy2112@gmail.com

**Arvind Balaji**
Texas A&M University
arvindb02@gmail.com

## Abstract

Prior work in Argument Mining frequently alludes to its potential applications in automatic debating systems. Despite this focus, almost no datasets or models exist which apply natural language processing techniques to problems found within competitive formal debate. To remedy this, we present the DebateSum dataset[1]. DebateSum consists of 187,386 unique pieces of evidence with corresponding argument and extractive summaries. DebateSum was made using data compiled by competitors within the National Speech and Debate Association over a 7-year period. We train several transformer summarization models to benchmark summarization performance on DebateSum. We also introduce a set of fasttext word-vectors trained on DebateSum called debate2vec. Finally, we present a search engine for this dataset which is utilized extensively by members of the National Speech and Debate Association today. The DebateSum search engine is available to the public here: http://www.debate.cards

## 1 Introduction and Background

American competitive debate's increasingly technical nature leads its competitors to search for hundreds of thousands of pieces of evidence every year. While there are many types of competitive debate that can be assisted by Argument Mining technologies, some types of formal debate are easier to automate than others. In the United States, the National Speech and Debate Association (NSDA) organizes the majority of competitive debates held in secular high schools. The NSDA sanctions four different types of debate and many different speaking events. The NSDA-sanctioned debate format most suited to being assisted by Natural Language Processing technologies is called "Cross Examination Debate" (CX) or "Policy Debate". This is because Policy Debate is significantly more popular and evidence-intensive than the other debate forms that the NSDA offers. Unlike other forms of debate, which have narrow topics that rotate per tournament or per month, Policy Debate maintains one extremely broad topic over a whole year. This encourages extremely deep and thorough amounts of preparation. Significantly more evidence (and subsequently, training data) is produced by NSDA Policy Debaters than by other types of debaters.

Most debaters are encouraged to keep their cases and evidence secretive and hidden from their opponents. However, due to the extreme research burdens which policy debaters face, many universities hold "debate camps" which students attend to prepare for the year's topic. The primary goal of debate camp is for attendees to produce as much evidence as they can before the competitive season starts. These debate camps attract thousands of coaches, competitors, and staff, and function as an effective crowed sourcing platform. At the end of the summer, these debate camps release all evidence gathered by the attendees together on the Open Evidence Project[2]. The Open Evidence Project hosts thousands of debate cases and hundreds of thousands documents. The Open Evidence Projects extensive case library gives any policy debater access to a wide variety of debate cases, allowing for novices and competitors with limited amounts of preparation time to present effective arguments.

The Open Evidence Project is a fantastic hidden-gem resource for argument mining. A diverse range of highly motivated high school students and their coaches produce high quality arguments and evidence to support each argument made. Policy debate does not focus on the speaking style or delivery of the speech as much as other types of debate do. Policy debates are instead extremely intricate and technical and most unexpectedly to lay-people, the debaters deliver them *fast*[3]. Since rounds are usually decided by technical details of evidence, competitors are encouraged to present the maximal amount of evidence for their position in their speeches. This evidence is always available to be reviewed later by the judge before a decision is made. This leads to a phenomenon known as speed-reading (colloquially referred to as "spreading" within the debate community) which is done by the majority of serious competitors for strategic benefits. To casual observers, spreading seems completely absurd, but the competitive advantages that it confers are significant. The desire to present as much evidence as possible motivates competitors to research extremely deeply and to produce/deliver the maximum amount of evidence possible. It is due to these factors that DebateSum is such a large dataset.

---

[1] The DebateSum dataset is available here: https://github.com/Hellisotherpeople/DebateSum
[2] The Open Evidence Project is hosted here: https://openev.debatecoaches.org/Main/

Figure 1: An example of an argument-evidence-summary triplet from the DebateSum dataset as presented in its original form before parsing. The argument is highlighted in blue (lines 1-4), and would be presented by debaters as part of their case. Metadata, such as the date, title and author of the evidence are highlighted in green (lines 5-9). The evidence consists of all text after the blue and green-highlighted sections. The extractive summary consists of all underlined text within the document. The highlighted sections of the underlined text are the extracts which the debater chooses to read out-loud alongside their argument. Note that the argument can also be used an abstractive summary or as a query in query-focused summarization

Figure 2: The debate.cards search engine. Debaters can quickly search for evidence by keywords. They can pick evidence to view in detail with the eyeball icon, or move it to the "saved" column with the arrow icon. When a debater has moved all of the evidence that they need to the "saved" column, they can click the download button (not shown) to download a properly formatted word document with all of the saved evidence in it. Policy Debaters extensively use this download feature before or in debate rounds to compile evidence for their case.

---

[3] It truly must be seen to be believed, an example of this can be found here, and it is the norm within the activity: `https://youtu.be/Q5iJ7mR0NNs?t=754`

This uniquely technical style of debate has thus far been largely unassisted by natural language processing technologies. Debaters would find that effective information retrieval, summarization (especially query-focused summarization), and classification systems could automate, supplement, and/or assist them in their research work. The primary reason for this lack of effective systems within competitive debate was the lack of domain specific training data available. DebateSum remedies this problem by introducing the largest (to our knowledge) dataset of document-argument-summary triplets in existence and we believe that it is one of the largest argument mining datasets ever gathered.

## 2 Innovations Introduced

We introduce three innovations: the DebateSum dataset (and summarization models trained on it), debate2vec word embeddings, and the "debate.cards" argument search engine.

Open Evidence stores tens of thousands of debate cases as Word documents. Some debate cases can have thousands of documents within them. DebateSum was gathered by converting each word document into an html5 file using pandoc[4]. This allows for easy parsing of documents, allowing for them to have their arguments, document, and summary extracted.

DebateSum consist of argument-evidence-summary triplets. Policy Debaters colloquially refer to these triplets as "cards". Each card consist of a debater's argument (which acts as a biased abstractive summary of the document), a supporting piece of evidence and its citations, and a word-level extractive summary produced by "underlining" and/or "highlighting" the evidence in such a way to support the argument being made. Figure 1 shows an example of one of these cards before parsing. Thousands of competitors and coaches manually produced and annotated DebateSum's triplets. We train transformer-based token-level extractive summarization models to form strong baselines and to act as a blueprint for performance benchmarking summarization models on DebateSum.

Debate2vec is a set of fasttext (Bojanowski et al., 2016) word vectors produced for study on word analogy and word similarity tasks. It is trained on the evidence in DebateSum along with additional evidence (which was not included in DebateSum due to missing or malformed arguments or summaries). Debate2vec is trained on 222485 documents. Debate2vec is 300 dimensional, with a vocabulary of 107555 lowercased words. It was trained for 100 epochs with a learning rate of 0.10. No subword information is included to keep the memory consumption of the model down. Debate2vec is particularly well suited for usage in domains related to philosophy, law, government, or politics.

Debate.cards is an argument search engine which indexes DebateSum (and additional evidence gathered by college level debaters or contributed by users later in the year). Debate.cards was designed as a research tool for Policy Debaters. Prior to debate.cards, competitors would have to painstakingly search through potentially thousands of word documents for specific pieces of evidence. Debate.cards allows competitors to search for evidence by keyword match within the argument, evidence or the citation. Figure 2 shows a sample of the debate.cards search engine.

## 3 Prior Work

We are not the first to utilize Natural Language Processing techniques to assist debaters with summarization. Abstractive summarization of political debates and arguments has been studied for quite some time (Egan et al., 2016), (Wang & Ling, 2016). The most well-known application of argument mining to debate comes from IBM's Project Debater. Recent work from IBM research has shown the impressive capabilities of automatic argument summarization (Bar-Haim et al., 2020). They utilize a crowd-sourced argument dataset of 7000 pro/con claims scraped from the internet and then they abstractivly synthesize key-points which summarizes the arguments made by the participants. Our work and dataset focuses instead on word-level extractive summarization of debate documents. There is work related to retrieving extractive "snippets" in support of an argument, but this work is query-independent and extracts summaries at the sentence level rather than the word level (Alshomary et al., 2020). Other work related to debate summarization exists, but is trained on limited datasets or restricted to significantly less technical debating styles and formats (Sanchan et al., 2017).

One notable feature that would be extremely useful to members of the Policy Debate community is the ability to generate an extractive summary of their evidence which is *biased towards supporting their*

---

[4] Available here: https://pandoc.org/

*argument*. Some evidence will make arguments for both sides, but only the portions which support a particular position would ideally be read aloud. Some authors have explored this unique problem which they call *query focused* or *query based* summarization, notably (Baumel et al., 2018), (Xu & Lapata, 2020) and (Nema et al., 2017). These systems are somewhat similar to our work, but they deal with abstractive rather than extractive summarization, and are trained on comparatively small datasets like the DUC 2007 or Debatepedia dataset. Luckily, a queryable word-level extractive summarization system exists and is used among the community[5]. This summarizer is called "CX_DB8" (to celebrate the "cross examination" debate community) and is unsupervised, which makes it unable to be directly trained on DebateSum (though it can use word embedding's which are trained on it). It also gets inferior ROUGE scores compared to supervised models. The summarization models we train on DebateSum are not queryable, but they follow the tradition of supervised extractive summarization models being benchmarked using ROUGE scores.

Before our work, there were (to our knowledge) no pre-trained word embedding's or language models publically available for Policy Debate. The closest pre-trained language model that we could find to our domain is the publically available Law2Vec[6] set of legal word embedding's. A significant proportion of policy debaters end up becoming lawyers, in no small part due to the similarity between competitive debate and litigation. Our Debate2Vec[7] word embedding's are trained on the entirety of the DebateSum document dataset. They are (to our knowledge) the largest publically available set of word vectors trained on a non-legal argumentative corpus

Finally, there is prior work related to argumentation search engines and information retrieval systems for debaters. Websites such as debate.org, procon.org, and debatepedia.org serve as useful search engines for arguments in support or opposition of a topic. Argument search engines which match or exceed the scale of DebateSum's arguments have been built by crawling these sorts of websites (Wachsmuth et al., 2018), but these indexed arguments do not have corresponding evidence and extracts associated with them. No dedicated search engine for competitive policy debate evidence existed prior to our work. Furthermore, we believe that no dedicated search engine for *any* type of debate argument-evidence-summary triplets exists that matches the scale or breadth of the Debate-Cards search engine.

## 4 Analysis

The DebateSum dataset consists of 187,386 unique documents that are larger than 100 words. There are 107,555 words which show up more than 10 times in the corpus. There are 101 million total words in DebateSum. Each document consists of on average 520 words. Each argument is on average 14 words long, and each summary consist of 198 words on average. The mean summary compression ratio is 0.46 and the mean argument compression ratio is 0.06.

DebateSum is made from evidence pertaining to each of the yearly policy debate resolutions. Since DebateSum consists of 7 years of evidence, there are 7 resolutions which it covers. Affirmative debate cases almost always advocate for a particular "plan" which implements a resolution. There are potentially an infinite number of topical plans. Negative teams can read a potentially infinite number of counter-plans or counter-advocacies alongside evidence for why the affirmative plan is bad. Debaters will be expected to prepare cases for both the affirmative and negative side and debate each side an equal number of times throughout a tournament. As a result, a considerable amount of evidence gathered in a particular year will be only tangentially related to the resolution as it must be generic enough to answer any type of plan or counterplan. There is a consistent "metagame" of popular and strong arguments which are crafted to be used for any topic on either side. Many of them have their roots in philosophical critiques of the plan or even the state itself which may appear to have little or no relevance to the resolution. One can get insights into the specific list of arguments made available by looking at the corresponding years page of Open Evidence and inspecting the DebateSum debate cases in their original form. A table which lists the official resolution and its year is presented below:

---

[5] Available here: https://github.com/Hellisotherpeople/CX_DB8
[6] Available here: https://archive.org/details/Law2Vec
[7] Available here: https://github.com/Hellisotherpeople/debate2vec

| YEAR | RESOLUTION |
|---|---|
| **2013-2014** | Resolved: The United States federal government should substantially increase its economic engagement toward Cuba, Mexico or Venezuela. |
| **2014-2015** | Resolved: The United States federal government should substantially increase its non-military exploration and/or development of the Earth's oceans. |
| **2015-2016** | Resolved: The United States federal government should substantially curtail its domestic surveillance. |
| **2016-2017** | Resolved: The United States federal government should substantially increase its economic and/or diplomatic engagement with the People's Republic of China. |
| **2017-2018** | Resolved: The United States federal government should substantially increase its funding and/or regulation of elementary and/or secondary education in the United States. |
| **2018-2019** | Resolved: The United States federal government should substantially reduce its restrictions on legal immigration to the United States. |
| **2019-2020** | Resolved: The United States federal government should substantially reduce Direct Commercial Sales and/or Foreign Military Sales of arms from the United States. |

## 5 Experiments and Result

We train transformer architecture neural network language models on word-level extractive summarization using the simple-transformers[8] framework. We formulate this problem as a token classification problem between "underlined" and "not-underlined" tokens. We fine-tune the existing weights for 5 epochs using early stopping and the adam (Kingma & Ba, 2015) optimizer. Fp16 is enabled for training. We evaluate our models on a test split of 18,738 documents. The ROUGE metric is used for measuring summarization quality. We evaluate using the default settings of py-rogue[9] on our models. We report the ROUGE F1 scores of these transformer models.

| **Model** | **ROUGE-1** | **ROUGE-2** | **ROUGE-L** |
|---|---|---|---|
| BERT-Large (Devlin et al., 2018) | 56.32 | 35.20 | 49.98 |
| GPT2-Medium (Radford et al., 2019) | 52.07 | 34.20 | 53.23 |
| **Longformer-Base-4096 (Beltagy et al., 2020)** | **60.21** | **38.53** | **57.21** |

Table 1: Comparison of different token classification transformer models fine-tuned on the training split of the DebateSum dataset.

## 6 Conclusion

In this paper, we presented a new large-scale argument mining and summarization dataset called DebateSum. Each row of DebateSum consists of a piece of debate evidence, a word-level extractive summary of that evidence, and the argument made with the evidence. The argument can also be used as an abstractive summary or as a query in tandem with the extract for query-focused extractive summarization. We also trained word vectors on the debate evidence subset of DebateSum. We showcased an innovative search engine for DebateSum which is extensively utilized by competitive debaters today. Finally, we fine-tuned several transformer models on word-level extractive summarization of DebateSum documents and measured their performance using the ROUGE metric. We observe that Longformer is superior to competitor models, likely because of the significantly larger sequence length allowing long-range context to assist in choosing which tokens to include in a summary. We release all code and data to the public.

---

[8] Found here: https://github.com/ThilinaRajapakse/simpletransformers
[9] Available here: https://github.com/Diego999/py-rouge